\documentclass{article}

\usepackage[final]{neurips_2024}

\usepackage[utf8]{inputenc} 
\usepackage[T1]{fontenc}    
\usepackage{hyperref}       
\usepackage{url}            
\usepackage{booktabs}       
\usepackage{amsfonts}       
\usepackage{nicefrac}       
\usepackage{microtype}      
\usepackage{xcolor}         
\usepackage{natbib}
\setcitestyle{numbers,square}

\usepackage{graphicx}
\usepackage{eqnarray}
\usepackage{amsmath} 

\usepackage{caption}
\usepackage{subfigure}

\title{Reference Trustable Decoding: A Training-Free Augmentation Paradigm for Large Language Models}

\author{Luohe Shi$^{1,\dagger}$, Yao Yao$^2$, Zuchao Li$^{1,\dagger}$\thanks{$\ $  Corresponding author. $^\dag$ Equal contribution.}, Lefei Zhang$^{1}$, and Hai Zhao$^{2}$ \\
$^{1}$National Engineering Research Center for Multimedia Software, \\
School of Computer Science, Wuhan University, Wuhan, 430072, P. R. China \\
$^{2}$Department of Computer Science and Engineering, Shanghai Jiao Tong University\\
{\tt shiluohe@whu.edu.cn, yaoyao27@sjtu.edu.cn, zcli-charlie@whu.edu.cn,}\\
{\tt zhaohai@cs.sjtu.edu.cn}\\
}

\begin{document}

\maketitle

\begin{figure}[h]
  \includegraphics[width=\textwidth]{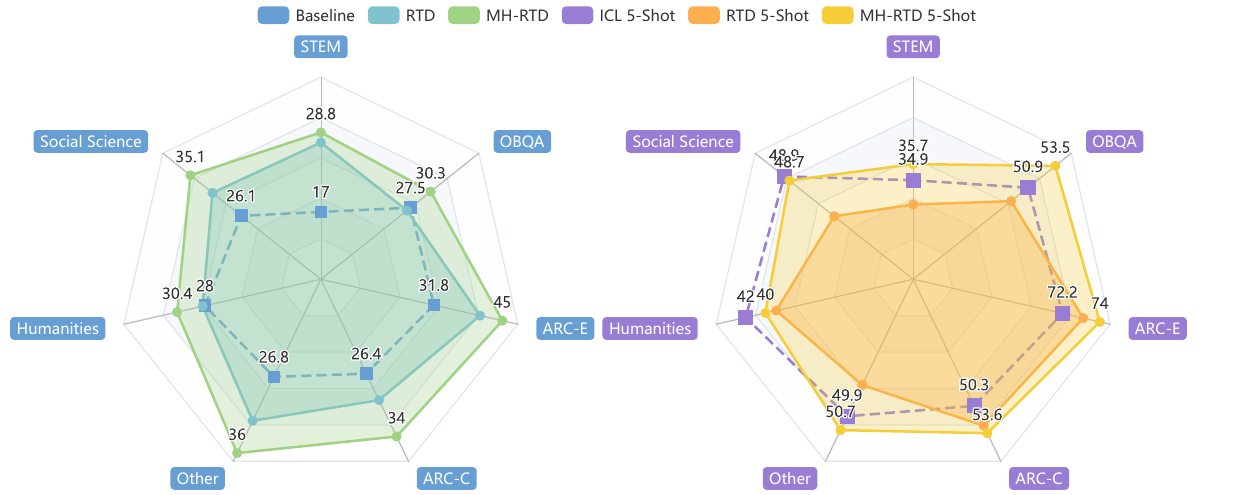}
  \caption{Performance comparison between default LLM and reference trustable decoding in reasoning tests.}
  \label{fig:figg1}
\end{figure}


\begin{abstract}
Large language models (LLMs) have rapidly advanced and demonstrated impressive capabilities. In-Context Learning (ICL) and Parameter-Efficient Fine-Tuning (PEFT) are currently two mainstream methods for augmenting LLMs to downstream tasks. ICL typically constructs a few-shot learning scenario, either manually or by setting up a Retrieval-Augmented Generation (RAG) system, helping models quickly grasp domain knowledge or question-answering patterns without changing model parameters. However, this approach involves trade-offs, such as slower inference speed and increased space occupancy. PEFT assists the model in adapting to tasks through minimal parameter modifications, but the training process still demands high hardware requirements, even with a small number of parameters involved.
To address these challenges, we propose Reference Trustable Decoding (RTD), a paradigm that allows models to quickly adapt to new tasks without fine-tuning, maintaining low inference costs. RTD constructs a reference datastore from the provided training examples and optimizes the LLM's final vocabulary distribution by flexibly selecting suitable references based on the input, resulting in more trustable responses and enabling the model to adapt to downstream tasks at a low cost.
Experimental evaluations on various LLMs using different benchmarks demonstrate that RTD establishes a new paradigm for augmenting models to downstream tasks. Furthermore, our method exhibits strong orthogonality with traditional methods, allowing for concurrent usage. Our code can be found at \href{https://github.com/ShiLuohe/ReferenceTrustableDecoding}{https://github.com/ShiLuohe/ReferenceTrustableDecoding}

\end{abstract}

\section{Introduction}

In the rapidly advancing field of artificial intelligence, Large Language Models (LLMs) have demonstrated substantial progress. With their extensive parameter size, LLMs have acquired emergent abilities~\cite{DBLP:journals/tmlr/WeiTBRZBYBZMCHVLDF22} and been able to tackle diverse and challenging tasks in fields like education~\cite{kasneci2023chatgpt} and medicine~\cite{thirunavukarasu2023large}. Despite their immense potential, Large Language Models that have just completed pre-training often struggle to effectively adapt to downstream tasks. Moreover, the process of adapting the model is typically costly and requires careful execution by experienced individuals. Otherwise, it could lead to the model generating hallucination~\cite{zhang2023sirens, DBLP:journals/corr/abs-2405-14744} at best, or at worst, result in a loss of its language capabilities.

\begin{figure*}
    \centering
    \includegraphics[width=1\linewidth]{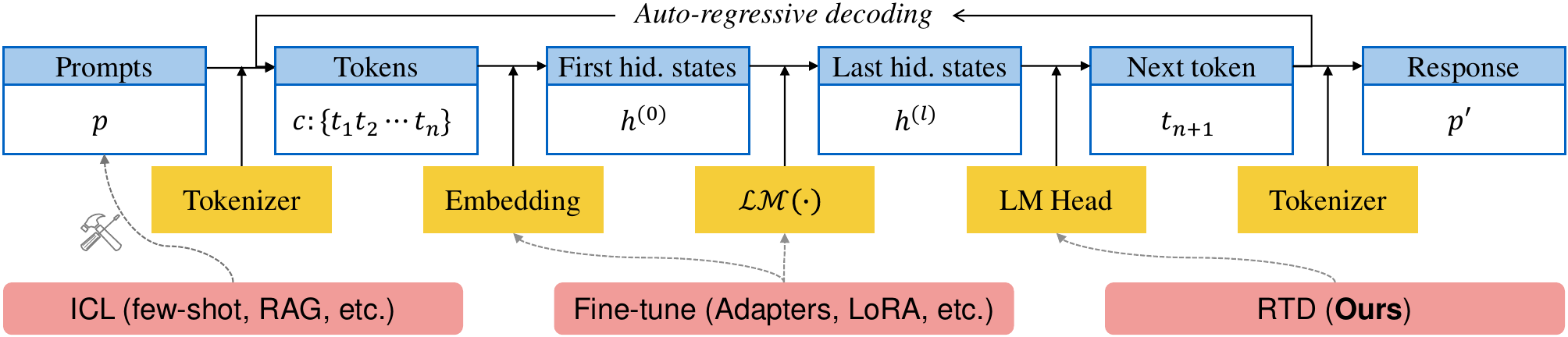}
    \caption{The pipeline of LLM inference and the focus of different methods: ICL focuses on the prompt stage, emphasizing the optimization of the model's input. Fine-tuning methods optimize the model itself by adjusting its parameters. In contrast, our proposed RTD method targets the decoding stage of the language model. By constructing a reference datastore, RTD optimizes the final output distribution without requiring additional training.}
    \label{fig:pipeline}
\end{figure*}

In-Context Learning (ICL), as a category of methods that do not require parameter adjustments, is one of the mainstream methods for adapting models to downstream tasks. ICL embeds domain knowledge, question-answering patterns, etc., into prompts through few-shot learning~\cite{NEURIPS2020_1457c0d6}, prompt engineering~\cite{zhou2023large}, and Retrieval-Augmented Generation (RAG)~\cite{liu2020retrieval} methods, leveraging the learning ability of the model itself to provide better answers. As pointed out in Figure~\ref{fig:pipeline}, ICL focuses on the prompt stage. However, ICL significantly increases the length of the input, consequently increases the space occupied by the KV-Cache required for inference. Further, according to the Roofline model~\cite{280922}, this part of the KV-Cache cannot be parallelized through batch processing, making memory I/O throughput a system bottleneck, wasting hardware computing power, and increasing token generation time during the entire inference stage.



Fine-tuning is also used to adapt models to downstream tasks. By fine-tuning the pre-trained model based on domain tasks, the model can quickly acquire capabilities within the domain. However, traditional full-parameter fine-tuning often requires a large amount of resources (empirically 8-15 times that of inference), making Parameter-Efficient Fine-Tuning (PEFT) a more popular method. By freezing most parameters and only modifying a few, methods such as Adapters, P-tuning~\cite{liu2022p}, LoRA~\cite{hu2021lora} and others~\cite{zhang-etal-2024-selective, DBLP:conf/icml/SongLZ0024, yang-etal-2024-soft} have become mainstream methods for quickly adapting models to downstream tasks. However, fine-tuning methods introduce several hyperparameters, which require high experience from the fine-tuners and the effects are unpredictable. Furthermore, due to the need for backpropagation, the computation graph must be saved, meaning that even if only a few parameters need to be updated, there will be a large amount of additional computation and space requirements (several times that of inference), raising the threshold for methods based on fine-tuning.

To address these challenges, we introduce Reference Trustable Decoding (RTD), a novel framework designed to fit LLMs for downstream tasks.
Distinct from a conventional \texttt{LM\_Head} module, RTD strategically retrieves relevant references from a pre-constructed datastore, guided by the final hidden states of the language model. This approach not only enhances the final output distribution by recalculating it with the similarity score of the retrieved references but also allows for the seamless integration of new knowledge or constraints into the response generation process without increasing the input length or using gradient descent.

RTD, distinctively training-free, emphasizes compact input lengths to expedite inference. RTD's effectiveness was rigorously tested using varied benchmarks focused on different tasks and a Wikipedia-based knowledge injection scenario. On these benchmarks, RTD achieved results comparable to traditional methods like PEFT and ICL, providing significant improvement. Additionally, we combined RTD with traditional methods, further enhancing the model's capabilities and demonstrating the good orthogonality of RTD with other approaches.

Our contribution includes:
\begin{itemize}
    \item We propose a new paradigm, called RTD, for fitting LLMs for downstream tasks. RTD is a training-free method that focused on the decoding stage of large language models (LLMs), as a alternation of \texttt{LM\_Head}. It helps LLMs to adapt to different tasks with different demands and provide trustable response.

    \item RTD has achieved performance comparable to, or even better than, ICL and PEFT across different benches, while maintaining the desirable properties of training free and not introducing additional input lengths. This demonstrates the potential of RTD as a new paradigm for LLMs to adapt to downstream tasks. Furthermore, RTD can be seamlessly integrated with other existing methods, such as in-context learning (ICL) and fine-tuning. The combination of RTD, ICL, and fine-tuning has the potential to achieve even higher performance.
\end{itemize}

\section{Background and Related Work}
In the field of NLPs, Transformer models have gained influence rapidly after it get original proposed in 2017. As larger scaled model been introduced, especially giant ones like GPT3 which has 175 Billion of parameters~\cite{floridi2020gpt}, the training process is getting more and more expensive, hence to fit the LLMs for downstream tasks.

\subsection{Fine-tuning}

\paragraph{Full Parameter Fine-tuning}
Full parameter fine-tuning refers to fully optimizing all the parameters of the model during the fine-tuning process. Full parameter fine-tuning has the advantage of allowing the model to adapt more closely to the specific task at hand, as well as injecting more information into the model. However, it also has the disadvantage of being the most computationally expensive and time-consuming, as it requires to manipulate all parameters of the model, with the modern optimizer like Adam~\cite{jais2019adam}, 8 to 15 times of more extra GPU memory is demanded comparing to inference empirically, resulting a must of multi-GPU server or even cross sever training.

\paragraph{Parameter Efficient Fine-tuning}
Parameter Efficient Fine-tune (PEFT), for example, LoRA~\cite{hu2021lora} and P-tuning~\cite{liu2022p}, is introduced to make fine-tune more reachable. By freezing most of the model parameters and only let a small amount of them accumulate gradient, the GPU memory and computation resource can be cut down by a large margin~\cite{he2021effectiveness}.

However, as fine-tuning introduce many tricky hyper-parameters like learning rate, the process is heavily task related and empirical, even experienced fine-tuner need some trials and error when tuning them.
Moreover, even if the number of parameters trained is not large, processes such as backpropagation still need to be carried out. The computation graph generated on long sequences will also occupy a large amount of memory, making the threshold for computing power and memory still high, which any method that relies on gradient descent is difficult to avoid.

\subsection{In-Context Learning}
\paragraph{Few-Shot Learning}
Few-shot Learning is proved to be a great way for LLMs to gain capability. By appending the true task that LLMs are expecting to response after a couple of existing correct examples, LLMs can gain its reasoning ability~\cite{NEURIPS2020_1457c0d6, nie2022improving}. 

\paragraph{Retrieval Augmented Generation}
Retrieval Augmented Generation (RAG) ~\cite{lewis2020retrieval} is an AI framework for retrieving facts from an external knowledge source to LLMs, which helps LLMs correct its hallucination and use latest fact~\cite{shuster2021retrieval}. RAG is to cut external knowledge source into multiple chunks, then embed and store them in a database, then retrieve them at the process of generation to let LLMs get the knowledge in it. This technique allowed LLM to use extra information while maintaining their parameters untouched. RAG have been used on multiple fields, like coding ~\cite{liu2020retrieval} and question answering~\cite{mao2020generation}. And it can be combined with few-shot~\cite{izacard2022few}.

The main drawback of ICL methods lies in their growth of the input sequence. Under the quadratic complexity of the Transformer architecture, this implies a longer KV-Cache, which not only increases the latency during the pre-fill stage but also adds delay each time a token is generated~\cite{yao-etal-2024-sirllm}. Moreover, unlike model parameters, each instance needs to save its dedicated portion of KV-Cache, leading to memory I/O bottlenecks and computational power waste. Finally, on some smaller models, the irrelevant information that ICL might contain can confuse the model, resulting in performance loss.

\section{Reference Trustable Decoding}
\label{sec:RTDMethod}
In this section, we begin by presenting the fundamental formulas and concepts to elucidate the workings of Reference Trustable Decoding, followed by an exploration of the multi-head Reference Trustable Decoding method.

\subsection{Preliminary}
Given an input sentence $c=\{t_1,t_2,...,t_{n}\}$, where $t_i$ represents the i-th token and $n$ denotes the sentence length, the last token's output of the last Transformer block in the language model can be represented as:
\begin{equation}
    h^{(l)} = \mathcal{LM}(c)
\end{equation}
In this equation, $h^{(l)} \in \mathbb{R}^{d_m}$ is the output of the last token from the final, or the $l$-th, Transformer block of the language model, where $d_m$ denotes the hidden size of the model. 

Traditionally, a standard decoder-only architecture Transformer usually employs \texttt{LM\_Head}, which is, a fully connected layer, usually includes a learnable weight matrix $W$ and no bias, followed by a softmax function $\mathbf{Softmax}(\cdot)$ to predict the output probability distribution \textbf{p} of the next token from the last hidden states:
\begin{equation}
    \mathbf{p} =\mathtt{LM\_Head}(h^{(l)}) = \mathbf{Softmax}(W\cdot h^{(l)})
\end{equation}
where $v$ is the vocabulary size and $W \in \mathbb{R}^{v \times d_m}$. 

However, traditional next token prediction does not support incorporating external information and therefore, we introduce reference trustable decoding where we build a bypass around the \texttt{LM\_Head}, showcased in Figure~\ref{fig:overview_rtd}, as the entrance of additional knowledge or guidance.

\subsection{Reference Trustable Decoding}
\subsubsection{Generation of Reference Datastore}

\begin{figure*}
    \centering
    \includegraphics[width=1\linewidth]{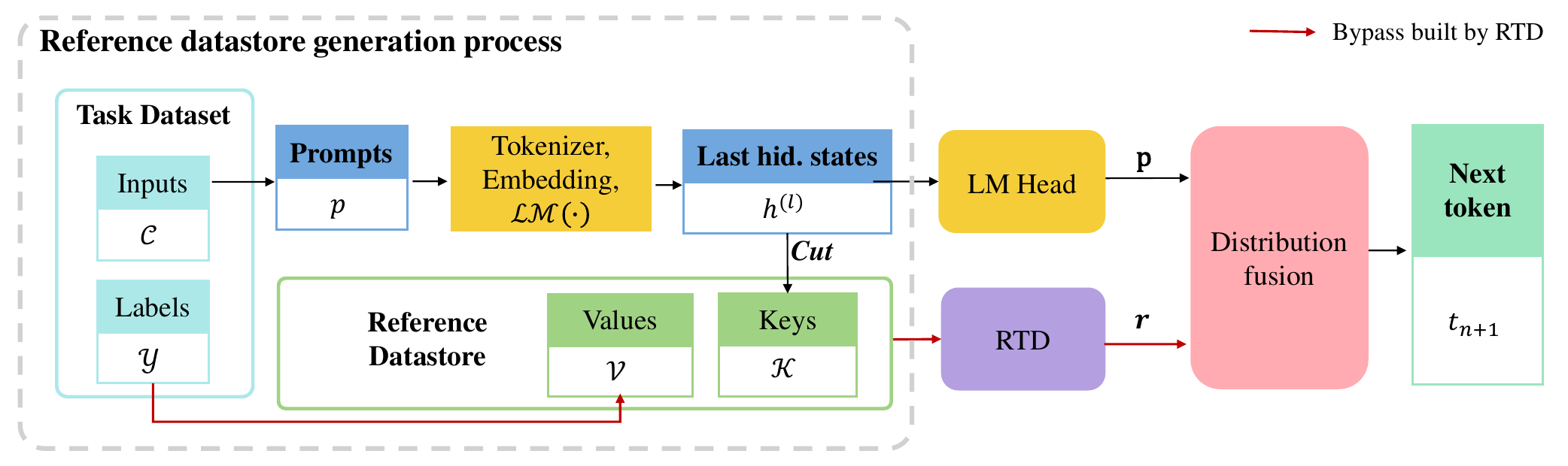}
    \caption{Overview of the reference datastore generation and reference trustable decoding process.}
    \label{fig:overview_rtd}
\end{figure*}

In reference trustable decoding, we first build the reference datastore $\mathcal{L}$, which stores key-value pairs $(k, v) \in (\mathcal{K}, \mathcal{V})$. Here, the key $k = \mathcal{LM}(c)$ represents the last hidden states of the token generated by the LMs from the context $c$, and the value $v$ is the corresponding label $y$. Mathematically, we have:

\begin{equation}
    \mathcal{L} = \{(k ,v)|(k ,v) \in (\mathcal{K}, \mathcal{V})\} 
    =\{\left(\mathcal{LM}\left(c\right), y\right)\ |\ (c, y) \in \mathcal{D}\}
\end{equation}
where $\mathcal{D}=(\mathcal{C},\mathcal{Y})$ is the task dataset with input context set $\mathcal{C}$ and label set $\mathcal{Y}$, and $|\mathcal{Y}|$ refers the number of possible labels. This process is depicted in Figure~\ref{fig:overview_rtd}. It's obvious that \textbf{the computational requirement is same as performing a forward pass to every content in the task dataset}, which aligned with the minimal requirement of the inference stage, denotes the superiority of RTD as a gradient-free method.



\subsubsection{Decoding Stage}
\begin{figure*}
    \centering
    \includegraphics[width=1\linewidth]{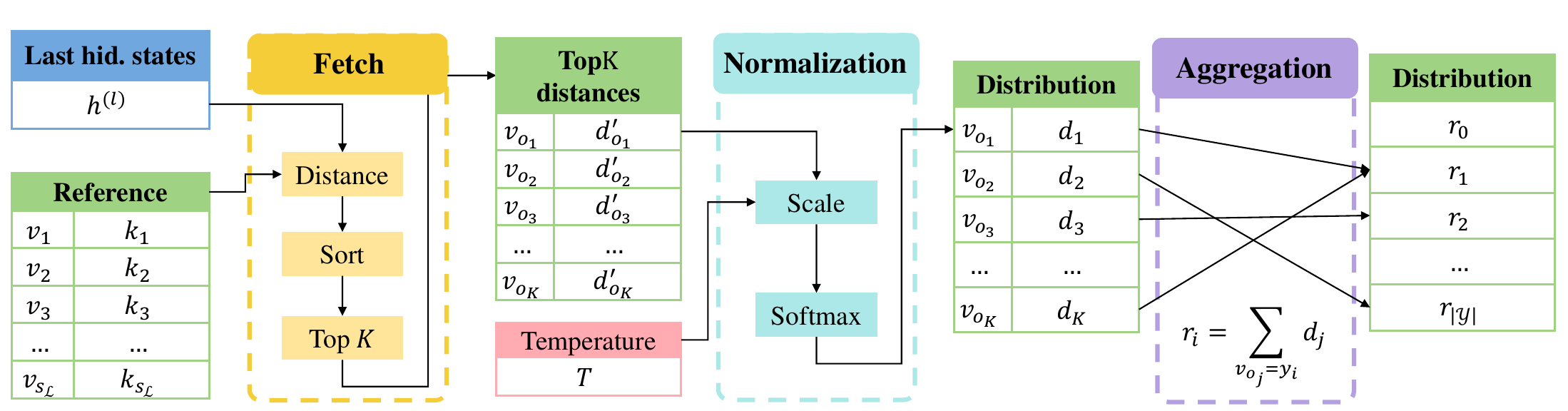}
    \caption{Three stages of reference trustable decoding.}
    \label{fig:g3}
\end{figure*}

At each decoding round, given the input context $c$, we first compute $h^{(l)} = \mathcal{LM}(c)$, which is the input to RTD and \texttt{LM\_Head}. Then we use a three stage approach to get the RTD output, \textbf{Fetch}, \textbf{Normalization}, and \textbf{Aggregation}, depicted in Figure~\ref{fig:g3}. 

\paragraph{Fetch} 
First, we calculate the distance $d'$ between $h^{(l)}$ and all the $k$ in the reference datastore $\mathcal{L}$. Otherwise stated, we use the Euclidean distance $d'_{i} = || h^{(l)} - k_i ||_2$. 
We then select the top $K$ instances from $\mathcal{L}$ which have the smallest distance, and for the $j$-th ($1\leq j\leq K$) closest $(k_i, v_i)$, we define $o_j=i$. Then we create a set $L_h$, storing the top $K$ distances and values:
\begin{equation}
    L_h = \{(d'_{o_j}, v_{o_j})\} = \{(|| h^{(l)} - k_{o_j} ||_2, v_{o_j})\}
    ,\quad
    o_j=i\ \mathrm{for}\ j\mathrm{-th\ closest\ }(k_i, v_i)  
\end{equation}

\paragraph{Normalization}
We first scale the $d'$ we got from the previous stage by temperature $T$, as $d''_{j}={d'_{o_j}}/{T}$. The scale operation is introduced to prevent overflow in the following Softmax operation. We take the Softmax of $-d''$ as $d$, guaranteed $d$ as a valid possibility distribution.
\begin{equation}
    d = \mathbf{Softmax}(-d''),\quad
    d_{j} = \frac{\mathrm{exp}\{-d''_{j}\}}{\sum_{\iota=1}^{K}{\mathrm{exp}\{-d''_{\iota}\}}}
    = \frac{\mathrm{exp}\{-d'_{o_j} / T\}}{\sum_{\iota=1}^{K}{\mathrm{exp}\{-d'_{o_\iota} / T\}}}
\end{equation}

\paragraph{Aggregation}
We calculate the final reference possibility distribution $\mathbf{r}=[r_1,r_2,...,r_{|\mathcal{Y}|}] \in \mathbb{R}^{|\mathcal{Y}|} $ by aggregating all $d_j$ that satisfies $v_{o_j} = y_i$, where $y_i\in \mathcal{Y}$.
\begin{equation}
    r_i = \sum_{v_{o_j} = y_i} d_j
\end{equation}

We denote $\mathcal{R}(\cdot, \mathcal{L}): \mathbb{R}^{d_m} \to \mathbb{R}^{|\mathcal{Y}|}$ as the function represents all three stages of querying the datastore $\mathcal{L}$ and building the corresponding reference possibility distribution $\mathbf{r}$. Therefore, we have 
\begin{equation}
    \mathbf{r} = \mathcal{R}(h^{(l)}, \mathcal{L})
\end{equation}

Additionally, when $|\mathcal{Y}|=v$, we can merge the distribution $\mathbf{p}$ given by $\mathtt{LM\_Head}(\cdot)$ and $\mathbf{r}$ given by $\mathcal{R}(\cdot, \mathcal{L})$ with a hyper-parameter $\lambda$:
\begin{equation}
    d' = \lambda \cdot \mathbf{r} + (1-\lambda) \cdot \mathbf{p}
\end{equation}
which is a common fusion method for mixing two distributions~\cite{see-etal-2017-get, Khandelwal2020Generalization, hashimoto-etal-2019-high, Gu_Wang_Cho_Li_2018}. 

\subsection{Multi-head Reference Trustable Decoding}
\begin{figure*}
    \centering
    \includegraphics[width=0.95\linewidth]{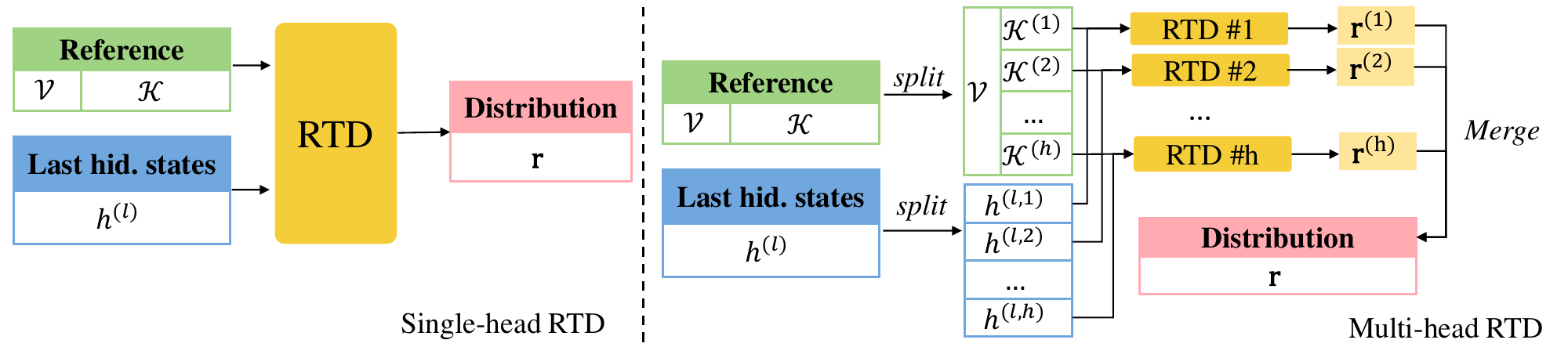}
    \caption{Comparison between RTD and multi-head RTD.}
    \label{fig:mhcmp}
\end{figure*}

Large language models like LLaMA2-70B~\cite{touvron2023llama} or Mistral-7B~\cite{Mistral7B} utilized MHA and GQA mechanism~\cite{AinslieLJZLS23}, implies the potential of splitting a large attention vector into smaller ones. So we adapt this method into our RTD process. We define $n_h$ of the head count of the LM model, and $d_h$ the dimension of the each attention head where $d_m = n_h \cdot d_h$. with this in mind, we split the reference datastore into $n_h$ sub-datastore by head. When decoding, we first split $h^{(l)}$ in to heads, then query each sub-datastore and merge the result, showcased in Figure~\ref{fig:mhcmp}. Mathematically,
\begin{equation}
\begin{aligned}
    &k^{(i)} = k \left[ d_h \times (i-1):\ d_h\times i\right],\quad
    h^{(l, i)} = h^{(l)} \left[ d_h \times (i-1):\  d_h\times i\right] \\
    &\mathcal{L}^{(i)} = \{(k^{(i)} ,v)|(k ,v) \in (\mathcal{K}, \mathcal{V})\}
\end{aligned}
\end{equation}
And we denote $\mathcal{R}_{\mathrm{MH}}(\cdot, \mathcal{L})$ as the function of the multi-head RTD query process, we have:
\begin{equation}
    \mathbf{r} = \mathcal{R}_{\mathrm{MH}}(h^{(l)}, \mathcal{L}) 
    = \frac1{n_h}\sum_{i=1}^{n_h}\mathcal{R}(h^{(l,i)},\mathcal{L}^{(i)})
\end{equation}

\begin{table}
    \centering
    \caption{Comparison of RTD and MH-RTD on Open Book QA.}
    \label{tab:t1}
    \begin{tabular}{l|cc}
        \toprule
        Method              & RTD         & MH-RTD\\
        \midrule
        MPT-7B              & 27.4       & 30.9 \\
        LLaMA2-7B           & 47.1       & 52.4 \\
        LLaMA2-70B          & 63.3       & 65.6 \\
        \bottomrule
    \end{tabular}
\end{table}

\subsection{Time and Memory Consumption}
\label{sec:effie}
\paragraph{Time Consumption}
\label{tab:tc}
The time consuming is largely depended on the vector datastore used. For a brute force searching datastore, the time complexity will be $\mathcal{O}\left(s_{\mathcal{L}}\cdot d_m\right)$ where $s_{\mathcal{L}} = |\mathcal{L}|$ is the size of the datastore. However, for those more powerful database like faiss~\cite{johnson2019billion} by Meta, with extra training after the generation of reference datastore, the process which have to be done again if the datastore changes, the time consumption can be cut to $\mathcal{O}\left(k\cdot d_m\right)$, where $k$ is a constant related the parameters used to train the database.

For multi-head reference trustable decoding, the performance cost remains the same. The time complexity of each attention-head wise query is $\mathcal{O}\left(d_{h}\cdot s_{\mathcal{L}}\right)$, the overall query time complexity is $\mathcal{O}\left(n \cdot d_{h}\cdot s_{\mathcal{L}}\right) = \mathcal{O}\left(d \cdot s_{\mathcal{L}}\right)$, which is the time complexity of convention reference trustable decoding processing. The calculation remains the same for a trained database, the overall time complexity is $\mathcal{O}\left(n \cdot k\cdot d_{h}\right) = \mathcal{O}\left(k\cdot d_m\right)$. 

\paragraph{Memory Consumption}
The use of time can be optimized by utilizing vector database, however the memory consumption cannot shrink easily. We further define $b$ as the bit cost of the models' $\mathtt{dtype}$, where $b_{\mathtt{float32}}=4, b_{\mathtt{float16}} =b_{\mathtt{bfloat16}}=2, b_{\mathtt{int8}}=1, b_{\mathtt{int4}}=\frac12$. The overall memory cost is $d_m \cdot b \cdot s_{\mathcal{L}}$. Due to the lack of lower precision $\mathtt{dtype}$ support on CPU, even the base model utilized popular half precision $\mathtt{dtype}$ like $\mathtt{bfloat16}$, it still need to be converted into larger ones to be stored. Since all the hidden states have to be saved to calculate precise distanced when rescalled, the memory cost can't be reduced significantly by making it irrelevant with $s_{\mathcal{L}}$. 

On the Multi-head RTD side, the memory cost remains the same as the regular RTD takes. The proof is same as the Section~\ref{tab:tc}. For instance, reference datastore and head-wise reference datastore with $20,480$ entries with $d_m = 4096$, $n = 32$ and $d_h = 128$, stored in $\mathtt{float32}$, takes $320\mathtt{MB}$ of memory and hard disk space.

\paragraph{MH-RTD for Resource Saving}
\label{sec:aht0}
As MH-RTD splits long vectors into multiple smaller ones, it gives us the opportunity to cut time and memory cost by merging different heads together, or directly evict some of them.
If on average, $p$ heads are merged into one head, then we expect a $\frac1p$ resource consumption. The time and memory improvement and corresponding performance impact can be find in the tuning Section~\ref{tab:aht}. 

\section{Settings and Experiment}
\label{sec:exper}
We categorize the common downstream tasks of language models into two types: language understanding and language generation. The former focuses on understanding the input information, based on the context and the information stored within the model, and then outputs the answer in the form of a few tokens, usually in a very simple form. The latter focuses on generating new sentences with complete semantics. We explored the potential of RTD compared to other methods on these two types of tasks. We first compared the effects of RTD and MH-RTD. As shown in Table~\ref{tab:t1}, we found that MH-RTD effectively enhances the capabilities of RTD. Therefore, we default to using the MH-RTD method in the following tests.

\subsection{Language Understanding}
We tested the language understanding capabilities of RTD on multiple benchmarks.
When testing, question without answer be shown to the LLM, then we will gather it's baseline output by LLMs' first output token and our RTD result through searching our reference datastore. $\lambda$ is set to $1$ in this task. How the reference datastore is generated can be fount at appendix~\ref{sec:gorl}.

Models we used are: LLaMA2-7B and 70B~\cite{touvron2023llama}, LLaMA3-8B~\cite{llama3modelcard}, MPT-7B~\cite{MosaicML2023Introducing}, GLM3-6B~\cite{zeng2022glm} ~\cite{du2022glm}, Yi-34B. Includes model size from 6B to 70B, as most of the major current models are. We use the \textit{base} version of the model by default. Testing benchmarks are: Massive Multitask Language Understanding (MMLU) ~\cite{hendrycks2020measuring}, AI2 Reasoning Challenge (ARC, both Easy (E) and Challenge (C) parts) ~\cite{allenai:arc}, Reasoning about Physical Commonsense in Natural Language (PIQA) ~\cite{Bisk2020}, Open Book Question Answering (OBQA) ~\cite{OpenBookQA2018}, and Massive Multitask Language Understanding in Chinese (CMMLU) ~\cite{li2023cmmlu}. C-MMLU is a Chinese benchmark, so only Chinese models, GLM3 and Yi, participated in this benchmark. 

The multiple-choice benchmarks we chose is challenging enough in itself and requires strong reasoning ability from the model; moreover, the answer format is fixed, which can simultaneously detect the ability to follow instructions. Since that most tasks in the traditional NLP field can be quickly converted into tasks of choosing one from several categories, even some generative tasks, so the results on the multiple-choice test can also represent many other tasks. 

The performance boost can be found both with or without ICL. Results are in table~\ref{tab:t2}. Besides testing scores, we also record the confused rate of baseline, the proportion of the questions that failed to be answered properly, including output irrelevant text or can't give a certain answer, in table~\ref{tab:tcf}. Meanwhile RTD is designed to given the LLMs' decision in a trustable and controllable way.
In comparison with fine-tuning methods in table~\ref{tab:t3}, we can notice that RTD can achieve approximate performance improvements as using PEFT methods like LoRA. Although it is still insufficient compared to full-parameter fine-tuning, the latter has a higher cost and has undergone knowledge injection (which is not considered in this part of the experiment). The dataset used for full-parameter fine-tuning is MMLU-Recall~\cite{LLaMA27BMMLU, MMLUrecall}, and the hyper-parameters of LoRA can be found in Appendix~\ref{sec:HPP}. Moreover, we've tested obqa score with different source of reference library, testing the generalization ability of RTD, as shown in Table~\ref{tab:genera}, RTD yields satisfactory results. We've also tested the performance of RTD with different $\lambda$ for language understanding, shown in Table~\ref{tab:lambd}. Lastly, we've tested the iteration speed of these benchmarks, as shown in Table~\ref{tab:timeeff}, the efficiency impact of RTD is minimized comparing to ICL.

\begin{table*}
    \centering
    \begin{tabular}{ll|cccc}
        \toprule
        Model & Benchmark   & Baseline & 5-shot ICL & RTD ($\Delta$)& 5-shot RTD ($\Delta$) \\
        \midrule
        LLaMA2-7B & MMLU        & 43.8           & 45.8            & 45.1 (1.3↑)& \textbf{47.2} (2.1↑)\\
        & ARC (E \& C)& 30.1           & 65.0            & 41.4 (11.3↑)& \textbf{67.3} (2.3↑)\\
        & PIQA        & 56.5           & 62.1            & 71.4 (14.9↑)& \textbf{73.2} (11.1↑)\\
        & Openbook QA & 27.8           & 51.0            & 30.4 (2.6↑)& \textbf{53.6} (2.6↑)\\
        \midrule
        LLaMA2-70B & MMLU        & 56.7           & 67.9            & 56.9 (0.2↑) & \textbf{68.5} (0.6↑) \\
        & ARC (E \& C)& 67.4           & 91.6            & 86.1 (19.7↑)& \textbf{91.7} (0.1↑) \\
        & PIQA        & 72.3           & 85.3            & 81.9 (9.6↑) &\textbf{86.6} (1.3↑)\\
        & OpenbookQA  & 53.7           & 84.4            & 68.2 (14.5↑)& \textbf{85.4} (1.0↑) \\
        \midrule
        LLaMA3-8B & MMLU & 47.5   & \textbf{63.9}  & 57.2 (9.7↑) & 61.9 (2.0↓)  \\
        & ARC (E \& C) & 71.2   & \textbf{87.3}  & 83.7 (12.5↑) & 87.1 (0.2↓) \\
        & PIQA        & 69.9           & 78.9            & 76.3 (6.4↑) &\textbf{80.0} (1.1↑)\\
        & OpenbookQA & 53.3   & 77.5  & 71.4(18.1↑) & \textbf{78.6} (1.1↑) \\
        \midrule
        MPT-7B &MMLU        & 27.4           & 29.6            & \textbf{30.4} (3.0↑) & 29.8 (0.2↑) \\
        & ARC (E \& C)& 27.5           & failed          & 27.6 (0.1↑) & \textbf{30.1}\\
        & OpenbookQA  & 29.4           & failed          & 27.2 (2.2↓) & \textbf{30.4}\\
        \midrule
        GLM3-6B & MMLU        & 41.9           & 48.6            & 47.6 (5.7↑) & \textbf{49.8} (1.2↑) \\
        & ARC (E \& C)& 59.1           & 75.3            & 75.0 (15.9↑)& \textbf{76.5} (1.2↑) \\
        & PIQA        & 66.8           & 73.6            & \textbf{75.9} (9.1↑) & 74.5 (0.9↑)\\
        & OpenbookQA  & 55.1           & 67.1            & 64.0 (8.9↑) & \textbf{68.8} (1.7↑) \\
        & C-MMLU      & 48.8           & 54.5            & 53.3 (4.5↑) & \textbf{54.7} (0.2↑)  \\
        \midrule
        Yi-34B & MMLU        & 68.6           &  \textbf{74.3}        & 70.3 (1.7↑) & 73.3 (1.0↓) \\
        & ARC (E \& C)& 93.3           & 94.0            & 90.7 (2.6↓) & \textbf{94.6} (0.6↑) \\
        & PIQA        & 88.3           & 83.5            & \textbf{88.4} (0.1↑) & 87.7 (4.2↑)\\
        & OpenbookQA  & 83.5           & \textbf{89.8}   & 88.4 (0.9↑) & 88.8 (1.0↓)     \\
        & C-MMLU      & 70.3           & 81.0            & 73.9 (3.6↑)  & \textbf{81.8} (0.8↑)\\
        \midrule
        \textbf{Avg} & - & 56.41 & 65.28 & 63.31 & \textbf{68.88} \\
        \bottomrule
    \end{tabular}
    \caption{RTD on language understanding benches. Baseline refers to zero-shot performance. ICL exceeds MPT-7B's $2048$ context window, with a 0 score result, recorded as failed in the table.}
    \label{tab:t2}
\end{table*}

\begin{table}[t]
\begin{minipage}[t]{0.5\linewidth}
    \centering
    \caption{Confused rate.}
    \label{tab:tcf}
    \begin{tabular}{l|ccc}
        \toprule
        Model       & Llama2-7B  & GLM3-6B & Yi-34B\\
        \midrule
        Rate         & 8.6\%      & 11.81\% & 0.44\%\\
        \bottomrule          
    \end{tabular}
\end{minipage}
\begin{minipage}[t]{0.49\linewidth}
    \centering
    \caption{RTD comparing with fine-tune methods.}
    \label{tab:t3}
    \begin{tabular}{l|cccc}
        \toprule
        Methods      & baseline  & LoRA  & FT & RTD\\
        \midrule
        Score        & 41.9  & 42.5   & 46.31 & 42.8\\
        \bottomrule          
    \end{tabular}
\end{minipage}
\end{table}
\begin{table}[t]
\begin{minipage}[t]{0.4\linewidth}
    \centering
    \caption{Generalization of RTD.}
    \label{tab:genera}
    \begin{tabular}{l|ccc}
        \toprule
        Source & OBQA  & ARC  & MMLU \\
        \midrule
        OBQA   & 71.4  & 71.4   & 71.2 \\
        \bottomrule          
    \end{tabular}
\end{minipage}
\begin{minipage}[t]{0.59\linewidth}
    \centering
    \caption{Different $\lambda$ in Language Understanding}
    \label{tab:lambd}
    \begin{tabular}{l|cccccc}
        \toprule
        $\lambda$ & 1    & 0.8  & 0.6  & 0.4  & 0.2  & 0  \\
        \midrule
        OBQA      & 71.4 & 68.0 & 67.0 & 66.8 & 66.6 & 53.3 \\
        \bottomrule          
    \end{tabular}
\end{minipage}
\end{table}
\begin{table}[h]
    \centering
    \caption{Efficiency of RTD.}
    \label{tab:timeeff}
    \begin{tabular}{l|cccc}
        \toprule
        Methods     & baseline  & RTD  & ICL & ICL + RTD \\
        \midrule
        Speed(it/s) & 25.1 & 23.6 & 7.90 & 7.85 \\
        Extra Memory Usage (MB)
                    & 0    & \~16 & \~37 & \~52 \\
        \bottomrule          
    \end{tabular}
\end{table}
\begin{table}[t]
\begin{minipage}[t]{0.49\linewidth}
    \centering
    \caption{Comparison of RTD and RAG using Wikipedia on LLaMA2-7B-Chat. }
    \label{tab:Wikipedia}
    \begin{tabular}{l|cc}
        \toprule
        LLaMA2-7B-Chat     & Acc           & Latency (ms)\\
        \midrule
        Baseline           & 39.0          & \textbf{42.5}\\
        Wiki RAG           & 29.0          &  > 200\\
        Wiki RTD            & \textbf{44.4} & 46.5\\
        \bottomrule
    \end{tabular}
\end{minipage}
\begin{minipage}[t]{0.49\linewidth}
\centering
    \caption{PPL of the fitted model on domain datasets.}
    \label{tab:fitter}
    \begin{tabular}{l|ccc}
        \toprule
        Dataset     & Baseline & LoRA   & RTD\\
        \midrule
        Tiny-S      & 1.6982   & 1.3710 & 1.4501\\
        \bottomrule
    \end{tabular}
\end{minipage}
\end{table}

\begin{figure}[h]
    \centering
    \includegraphics[width=1\linewidth]{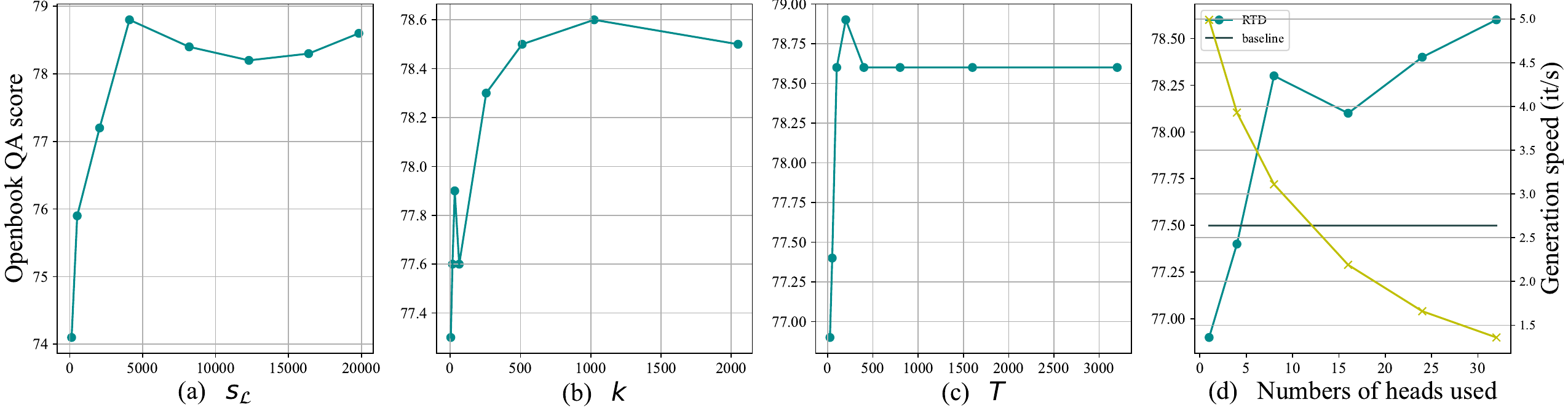}
    \caption{Hyper-parameters' influence on RTD's performance}
    \label{fig:abl}
\end{figure}

\subsection{Language Generation}

\paragraph{Reasoning with Context}
Generative tasks are generally subjective and difficult to test. We constructed a benchmark based on Retrieval-Augmented Generation (RAG) and Open Book Question Answering~\cite{OpenBookQA2018} to test the potential of RTD in areas requires advance reasoning such as knowledge injection. Chain-of-Thought~\cite{NEURIPS2022_9d560961} is a method that encourage the model to provide a step-by-step analysis before giving the final answer, thereby enhancing the model’s capabilities. We compared the performance of the model when introducing references through the ICL method and the RTD method, to determine the effectiveness of the RTD method. The extra knowledge source was Wikipedia. The generation of the datastore can be found in detailed in Appendix~\ref{sec:goll}. With the results of table \ref{tab:Wikipedia}, it can be seen that RTD was indeed helpful in knowledge injection. Besides, the context length is shrunk by a lot, thus saves reasoning GPU time and memory consumption. A detailed exploration of why RAG score is lower than baseline can be find in Appendix~\ref{sec:ragdef}.

\paragraph{Style transfer}
To explore whether the RTD method can be used to modify the language style of the model, we designed a style transfer experiment. We used a moderately scaled and strongly styled dataset, Tiny-Shakespeare~\cite{tinyssp, tinyssp2}, and compared the perplexity (PPL) of the model on the test set after LoRA and RTD, to measure whether our method can help the model change the output style. The results in Table~\ref{tab:fitter} prove that our RTD method can reduce the perplexity of the model, enabling the model to adapt to the style of different datasets. 
The hyperparameters of LoRA are in Appendix~\ref{sec:HPP}.

\subsection{Influence of Hyper-parameters in RTD}
\label{sec:ihpr}
Although our method is quick and efficient, it still introduces several hyper-parameters. We hope to explore the relationship between these hyper-parameters and the final performance of RTD. We conducted a series of ablation experiments on LLaMA2-7B~\cite{touvron2023llama} and OBQA~\cite{OpenBookQA2018} to explore the impact of different hyper-parameters on performance and how to quickly determine the optimal hyper-parameters. The overall result can be found in Figure~\ref{fig:abl}.
If not tuned, we set $k = 1024$, $s_{\mathcal{L}}=19,828$, $\lambda = 1$ and $T = 750$ by default.

Depicted in Figure~\ref{fig:abl} (a), RTD's performance improves initially with increasing $s_{\mathcal{L}}$ but eventually maxed out and starts oscillating when $s_{\mathcal{L}}$ reaches $4096$. Generally, a larger $s_{\mathcal{L}}$ gives a better performance, but it do get maxed out depends on the specific task.
%
Figure~\ref{fig:abl} (b) showcased us how RTD's performance consistently improves as $k$ increases initially, but eventually reaches a plateau, similiar with the $s_{\mathcal{L}}$. To be denoted is that a larger $k$ could harm efficiency.
%
Figure~\ref{fig:abl} (c) implies that RTD can only reach it's best performance when $T$ is large enough. Empirically, due to the characteristics of the exponential function, as long as the range of scaled distances $d''$ is kept between $1$-$2$, a sufficiently good effect can be achieved.
%
In RTD, $\lambda$ is an important variable, especially in generation tasks. However, $\lambda$ does not require high precision, and the range is relatively limited, so a good enough effect can be achieved quickly through a few attempts. Empirically speaking, $0.4$-$0.7$ is a suitable range for $\lambda$.
%
\label{tab:aht}
Previous studies indicated that by pruning the dimension of attention won't hurt $k$nn algorithm's performance~\cite{HeNB21}. In the case of RTD, showcased in Figure~\ref{fig:abl} (d), it can be found that the performance won't drop with at least $\frac14$ heads remained, and the generation speed was boosted as more heads are dropped.

\section{Conclusions}

In this paper, we introduce Reference Trustable Decoding, a novel training-free method designed to augment Large Language Models in downstream tasks. RTD refines the output distribution by leveraging references retrieved from a specially curated datastore, as a bypass of conventional \texttt{LM\_Head}. Our experimental results demonstrate RTD achieved superior performance compared to the In-Context Learning baseline in 21 out of 25 different dataset and model configurations as well as fine-tune based methods. This result highlights the effectiveness of RTD across a diverse range of scenarios, underscoring its potential as a robust solution for enhancing language model capabilities in downstream tasks.

\section*{Limitations \& Future Work}
\label{sec:limits}
RTD is an efficient and quick method to augment the capabilities of models on specific downstream tasks. However, for some tasks, especially generative tasks, the large reference datastores that are difficult to directly compress may pose challenges for applications. Nevertheless, we believe that there is likely inherent redundancy in such large datastores. We hope to enable machines to identify these redundancies while maintaining a gradient-free method, in order to achieve efficient fine-tuning. How to make RTD accomplish tasks with high quality while being space-efficient is our following research direction.

\section*{Acknowledgments}
We sincerely appreciate the valuable feedback provided by all reviewers during the review process, as well as the efforts of the area chairs.
This work was supported by the National Natural Science Foundation of China (No. 62306216), the Natural Science Foundation of Hubei Province of China (No. 2023AFB816), the Fundamental Research Funds for the Central Universities (No. 2042023kf0133).

{
\bibliography{neurips_2024}

\begin{thebibliography}{10}

\bibitem{Mistral7B}
Mistral AI.
\newblock Mistral 7b, the best 7b model to date, apache 2.0, 2023.

\bibitem{AinslieLJZLS23}
Joshua Ainslie, James Lee{-}Thorp, Michiel de~Jong, Yury Zemlyanskiy, Federico Lebr{\'{o}}n, and Sumit Sanghai.
\newblock {GQA:} training generalized multi-query transformer models from multi-head checkpoints.
\newblock In Houda Bouamor, Juan Pino, and Kalika Bali, editors, {\em Proceedings of the 2023 Conference on Empirical Methods in Natural Language Processing, {EMNLP} 2023, Singapore, December 6-10, 2023}, pages 4895--4901. Association for Computational Linguistics, 2023.

\bibitem{bandarkar2023belebele}
Lucas Bandarkar, Davis Liang, Benjamin Muller, Mikel Artetxe, Satya~Narayan Shukla, Donald Husa, Naman Goyal, Abhinandan Krishnan, Luke Zettlemoyer, and Madian Khabsa.
\newblock The belebele benchmark: a parallel reading comprehension dataset in 122 language variants.
\newblock {\em arXiv preprint arXiv:2308.16884}, 2023.

\bibitem{allenai:arc}
Sumithra Bhakthavatsalam, Daniel Khashabi, Tushar Khot, Bhavana~Dalvi Mishra, Kyle Richardson, Ashish Sabharwal, Carissa Schoenick, Oyvind Tafjord, and Peter Clark.
\newblock Think you have solved direct-answer question answering? try arc-da, the direct-answer {AI2} reasoning challenge.
\newblock {\em CoRR}, abs/2102.03315, 2021.

\bibitem{Bisk2020}
Yonatan Bisk, Rowan Zellers, Ronan~Le Bras, Jianfeng Gao, and Yejin Choi.
\newblock Piqa: Reasoning about physical commonsense in natural language.
\newblock In {\em Thirty-Fourth AAAI Conference on Artificial Intelligence}, 2020.

\bibitem{NEURIPS2020_1457c0d6}
Tom Brown, Benjamin Mann, Nick Ryder, Melanie Subbiah, Jared~D Kaplan, Prafulla Dhariwal, Arvind Neelakantan, Pranav Shyam, Girish Sastry, Amanda Askell, Sandhini Agarwal, Ariel Herbert-Voss, Gretchen Krueger, Tom Henighan, Rewon Child, Aditya Ramesh, Daniel Ziegler, Jeffrey Wu, Clemens Winter, Chris Hesse, Mark Chen, Eric Sigler, Mateusz Litwin, Scott Gray, Benjamin Chess, Jack Clark, Christopher Berner, Sam McCandlish, Alec Radford, Ilya Sutskever, and Dario Amodei.
\newblock Language models are few-shot learners.
\newblock In H.~Larochelle, M.~Ranzato, R.~Hadsell, M.F. Balcan, and H.~Lin, editors, {\em Advances in Neural Information Processing Systems}, volume~33, pages 1877--1901. Curran Associates, Inc., 2020.

\bibitem{du2022glm}
Zhengxiao Du, Yujie Qian, Xiao Liu, Ming Ding, Jiezhong Qiu, Zhilin Yang, and Jie Tang.
\newblock Glm: General language model pretraining with autoregressive blank infilling.
\newblock In {\em Proceedings of the 60th Annual Meeting of the Association for Computational Linguistics (Volume 1: Long Papers)}, pages 320--335, 2022.

\bibitem{llama3modelcard}
Abhimanyu Dubey, Abhinav Jauhri, Abhinav Pandey, Abhishek Kadian, Ahmad Al{-}Dahle, Aiesha Letman, Akhil Mathur, Alan Schelten, Amy Yang, Angela Fan, Anirudh Goyal, Anthony Hartshorn, Aobo Yang, Archi Mitra, Archie Sravankumar, Artem Korenev, Arthur Hinsvark, Arun Rao, Aston Zhang, Aur{\'{e}}lien Rodriguez, Austen Gregerson, Ava Spataru, Baptiste Rozi{\`{e}}re, Bethany Biron, Binh Tang, Bobbie Chern, Charlotte Caucheteux, Chaya Nayak, Chloe Bi, Chris Marra, Chris McConnell, Christian Keller, Christophe Touret, Chunyang Wu, Corinne Wong, Cristian~Canton Ferrer, Cyrus Nikolaidis, Damien Allonsius, Daniel Song, Danielle Pintz, Danny Livshits, David Esiobu, Dhruv Choudhary, Dhruv Mahajan, Diego Garcia{-}Olano, Diego Perino, Dieuwke Hupkes, Egor Lakomkin, Ehab AlBadawy, Elina Lobanova, Emily Dinan, Eric~Michael Smith, Filip Radenovic, Frank Zhang, Gabriel Synnaeve, Gabrielle Lee, Georgia~Lewis Anderson, Graeme Nail, Gr{\'{e}}goire Mialon, Guan Pang, Guillem Cucurell, Hailey Nguyen, Hannah Korevaar, Hu~Xu, Hugo
  Touvron, Iliyan Zarov, Imanol~Arrieta Ibarra, Isabel~M. Kloumann, Ishan Misra, Ivan Evtimov, Jade Copet, Jaewon Lee, Jan Geffert, Jana Vranes, Jason Park, Jay Mahadeokar, Jeet Shah, Jelmer van~der Linde, Jennifer Billock, Jenny Hong, Jenya Lee, Jeremy Fu, Jianfeng Chi, Jianyu Huang, Jiawen Liu, Jie Wang, Jiecao Yu, Joanna Bitton, Joe Spisak, Jongsoo Park, Joseph Rocca, Joshua Johnstun, Joshua Saxe, Junteng Jia, Kalyan~Vasuden Alwala, Kartikeya Upasani, Kate Plawiak, Ke~Li, Kenneth Heafield, Kevin Stone, and et~al.
\newblock The llama 3 herd of models.
\newblock {\em CoRR}, abs/2407.21783, 2024.

\bibitem{floridi2020gpt}
Luciano Floridi and Massimo Chiriatti.
\newblock Gpt-3: Its nature, scope, limits, and consequences.
\newblock {\em Minds and Machines}, 30:681--694, 2020.

\bibitem{Gu_Wang_Cho_Li_2018}
Jiatao Gu, Yong Wang, Kyunghyun Cho, and Victor~O.K. Li.
\newblock Search engine guided neural machine translation.
\newblock {\em Proceedings of the AAAI Conference on Artificial Intelligence}, 32(1), Apr. 2018.

\bibitem{günther2023jina}
Michael Günther, Jackmin Ong, Isabelle Mohr, Alaeddine Abdessalem, Tanguy Abel, Mohammad~Kalim Akram, Susana Guzman, Georgios Mastrapas, Saba Sturua, Bo~Wang, Maximilian Werk, Nan Wang, and Han Xiao.
\newblock Jina embeddings 2: 8192-token general-purpose text embeddings for long documents, 2023.

\bibitem{hashimoto-etal-2019-high}
Kazuma Hashimoto, Raffaella Buschiazzo, James Bradbury, Teresa Marshall, Richard Socher, and Caiming Xiong.
\newblock A high-quality multilingual dataset for structured documentation translation.
\newblock In {\em Proceedings of the Fourth Conference on Machine Translation (Volume 1: Research Papers)}, pages 116--127, Florence, Italy, August 2019. Association for Computational Linguistics.

\bibitem{HeNB21}
Junxian He, Graham Neubig, and Taylor Berg{-}Kirkpatrick.
\newblock Efficient nearest neighbor language models.
\newblock In Marie{-}Francine Moens, Xuanjing Huang, Lucia Specia, and Scott~Wen{-}tau Yih, editors, {\em Proceedings of the 2021 Conference on Empirical Methods in Natural Language Processing, {EMNLP} 2021, Virtual Event / Punta Cana, Dominican Republic, 7-11 November, 2021}, pages 5703--5714. Association for Computational Linguistics, 2021.

\bibitem{he2021effectiveness}
Ruidan He, Linlin Liu, Hai Ye, Qingyu Tan, Bosheng Ding, Liying Cheng, Jia{-}Wei Low, Lidong Bing, and Luo Si.
\newblock On the effectiveness of adapter-based tuning for pretrained language model adaptation.
\newblock In Chengqing Zong, Fei Xia, Wenjie Li, and Roberto Navigli, editors, {\em Proceedings of the 59th Annual Meeting of the Association for Computational Linguistics and the 11th International Joint Conference on Natural Language Processing, {ACL/IJCNLP} 2021, (Volume 1: Long Papers), Virtual Event, August 1-6, 2021}, pages 2208--2222. Association for Computational Linguistics, 2021.

\bibitem{hendrycks2020measuring}
Dan Hendrycks, Collin Burns, Steven Basart, Andy Zou, Mantas Mazeika, Dawn Song, and Jacob Steinhardt.
\newblock Measuring massive multitask language understanding.
\newblock In {\em 9th International Conference on Learning Representations, {ICLR} 2021, Virtual Event, Austria, May 3-7, 2021}. OpenReview.net, 2021.

\bibitem{hu2021lora}
Edward~J. Hu, Yelong Shen, Phillip Wallis, Zeyuan Allen{-}Zhu, Yuanzhi Li, Shean Wang, Lu~Wang, and Weizhu Chen.
\newblock Lora: Low-rank adaptation of large language models.
\newblock In {\em The Tenth International Conference on Learning Representations, {ICLR} 2022, Virtual Event, April 25-29, 2022}. OpenReview.net, 2022.

\bibitem{huang2023ceval}
Yuzhen Huang, Yuzhuo Bai, Zhihao Zhu, Junlei Zhang, Jinghan Zhang, Tangjun Su, Junteng Liu, Chuancheng Lv, Yikai Zhang, Jiayi Lei, Yao Fu, Maosong Sun, and Junxian He.
\newblock C-eval: A multi-level multi-discipline chinese evaluation suite for foundation models.
\newblock In {\em Advances in Neural Information Processing Systems}, 2023.

\bibitem{izacard2022few}
Gautier Izacard, Patrick S.~H. Lewis, Maria Lomeli, Lucas Hosseini, Fabio Petroni, Timo Schick, Jane Dwivedi{-}Yu, Armand Joulin, Sebastian Riedel, and Edouard Grave.
\newblock Atlas: Few-shot learning with retrieval augmented language models.
\newblock {\em J. Mach. Learn. Res.}, 24:251:1--251:43, 2023.

\bibitem{jais2019adam}
Imran Khan~Mohd Jais, Amelia~Ritahani Ismail, and Syed~Qamrun Nisa.
\newblock Adam optimization algorithm for wide and deep neural network.
\newblock {\em Knowledge Engineering and Data Science}, 2(1):41--46, 2019.

\bibitem{johnson2019billion}
Jeff Johnson, Matthijs Douze, and Herv{\'e} J{\'e}gou.
\newblock Billion-scale similarity search with {GPUs}.
\newblock {\em IEEE Transactions on Big Data}, 7(3):535--547, 2019.

\bibitem{kasneci2023chatgpt}
Enkelejda Kasneci, Kathrin Se{\ss}ler, Stefan K{\"u}chemann, Maria Bannert, Daryna Dementieva, Frank Fischer, Urs Gasser, Georg Groh, Stephan G{\"u}nnemann, Eyke H{\"u}llermeier, et~al.
\newblock Chatgpt for good? on opportunities and challenges of large language models for education.
\newblock {\em Learning and individual differences}, 103:102274, 2023.

\bibitem{Khandelwal2020Generalization}
Urvashi Khandelwal, Omer Levy, Dan Jurafsky, Luke Zettlemoyer, and Mike Lewis.
\newblock Generalization through memorization: Nearest neighbor language models.
\newblock In {\em International Conference on Learning Representations}, 2020.

\bibitem{lewis2020retrieval}
Patrick S.~H. Lewis, Ethan Perez, Aleksandra Piktus, Fabio Petroni, Vladimir Karpukhin, Naman Goyal, Heinrich K{\"{u}}ttler, Mike Lewis, Wen{-}tau Yih, Tim Rockt{\"{a}}schel, Sebastian Riedel, and Douwe Kiela.
\newblock Retrieval-augmented generation for knowledge-intensive {NLP} tasks.
\newblock In Hugo Larochelle, Marc'Aurelio Ranzato, Raia Hadsell, Maria{-}Florina Balcan, and Hsuan{-}Tien Lin, editors, {\em Advances in Neural Information Processing Systems 33: Annual Conference on Neural Information Processing Systems 2020, NeurIPS 2020, December 6-12, 2020, virtual}, 2020.

\bibitem{li2023cmmlu}
Haonan Li, Yixuan Zhang, Fajri Koto, Yifei Yang, Hai Zhao, Yeyun Gong, Nan Duan, and Timothy Baldwin.
\newblock {CMMLU:} measuring massive multitask language understanding in chinese.
\newblock {\em CoRR}, abs/2306.09212, 2023.

\bibitem{liu2020retrieval}
Shangqing Liu, Yu~Chen, Xiaofei Xie, Jing~Kai Siow, and Yang Liu.
\newblock Retrieval-augmented generation for code summarization via hybrid {GNN}.
\newblock In {\em 9th International Conference on Learning Representations, {ICLR} 2021, Virtual Event, Austria, May 3-7, 2021}. OpenReview.net, 2021.

\bibitem{liu2022p}
Xiao Liu, Kaixuan Ji, Yicheng Fu, Weng Tam, Zhengxiao Du, Zhilin Yang, and Jie Tang.
\newblock P-tuning: Prompt tuning can be comparable to fine-tuning across scales and tasks.
\newblock In {\em Proceedings of the 60th Annual Meeting of the Association for Computational Linguistics (Volume 2: Short Papers)}, pages 61--68, 2022.

\bibitem{DBLP:journals/corr/abs-2405-14744}
Xuan Liu, Jie Zhang, Song Guo, Haoyang Shang, Chengxu Yang, and Quanyan Zhu.
\newblock Exploring prosocial irrationality for {LLM} agents: {A} social cognition view.
\newblock {\em CoRR}, abs/2405.14744, 2024.

\bibitem{mao2020generation}
Yuning Mao, Pengcheng He, Xiaodong Liu, Yelong Shen, Jianfeng Gao, Jiawei Han, and Weizhu Chen.
\newblock Generation-augmented retrieval for open-domain question answering.
\newblock In Chengqing Zong, Fei Xia, Wenjie Li, and Roberto Navigli, editors, {\em Proceedings of the 59th Annual Meeting of the Association for Computational Linguistics and the 11th International Joint Conference on Natural Language Processing, {ACL/IJCNLP} 2021, (Volume 1: Long Papers), Virtual Event, August 1-6, 2021}, pages 4089--4100. Association for Computational Linguistics, 2021.

\bibitem{OpenBookQA2018}
Todor Mihaylov, Peter Clark, Tushar Khot, and Ashish Sabharwal.
\newblock Can a suit of armor conduct electricity? a new dataset for open book question answering.
\newblock In {\em EMNLP}, 2018.

\bibitem{nie2022improving}
Feng Nie, Meixi Chen, Zhirui Zhang, and Xu~Cheng.
\newblock Improving few-shot performance of language models via nearest neighbor calibration.
\newblock {\em arXiv preprint arXiv:2212.02216}, 2022.

\bibitem{LLaMA27BMMLU}
Liu Peng.
\newblock llama2-7b-mmlu, 2023.

\bibitem{MMLUrecall}
Liu Peng.
\newblock Mmlu-recall, employ mmlu (cmmlu) questions as initial seeds to retrieve related articles from multiple training data corpora., 2023.

\bibitem{see-etal-2017-get}
Abigail See, Peter~J. Liu, and Christopher~D. Manning.
\newblock Get to the point: Summarization with pointer-generator networks.
\newblock In {\em Proceedings of the 55th Annual Meeting of the Association for Computational Linguistics (Volume 1: Long Papers)}, pages 1073--1083, Vancouver, Canada, July 2017. Association for Computational Linguistics.

\bibitem{shuster2021retrieval}
Kurt Shuster, Spencer Poff, Moya Chen, Douwe Kiela, and Jason Weston.
\newblock Retrieval augmentation reduces hallucination in conversation.
\newblock In Marie{-}Francine Moens, Xuanjing Huang, Lucia Specia, and Scott~Wen{-}tau Yih, editors, {\em Findings of the Association for Computational Linguistics: {EMNLP} 2021, Virtual Event / Punta Cana, Dominican Republic, 16-20 November, 2021}, pages 3784--3803. Association for Computational Linguistics, 2021.

\bibitem{DBLP:conf/icml/SongLZ0024}
Weixi Song, Zuchao Li, Lefei Zhang, Hai Zhao, and Bo~Du.
\newblock Sparse is enough in fine-tuning pre-trained large language models.
\newblock In {\em Forty-first International Conference on Machine Learning, {ICML} 2024, Vienna, Austria, July 21-27, 2024}. OpenReview.net, 2024.

\bibitem{MosaicML2023Introducing}
MosaicML~NLP Team.
\newblock Introducing mpt-7b: A new standard for open-source, commercially usable llms, 2023.
\newblock Accessed: 2023-05-05.

\bibitem{thirunavukarasu2023large}
Arun~James Thirunavukarasu, Darren Shu~Jeng Ting, Kabilan Elangovan, Laura Gutierrez, Ting~Fang Tan, and Daniel Shu~Wei Ting.
\newblock Large language models in medicine.
\newblock {\em Nature medicine}, 29(8):1930--1940, 2023.

\bibitem{touvron2023llama}
Hugo Touvron, Louis Martin, Kevin Stone, Peter Albert, Amjad Almahairi, Yasmine Babaei, Nikolay Bashlykov, Soumya Batra, Prajjwal Bhargava, Shruti Bhosale, et~al.
\newblock Llama 2: Open foundation and fine-tuned chat models.
\newblock {\em arXiv preprint arXiv:2307.09288}, 2023.

\bibitem{DBLP:journals/tmlr/WeiTBRZBYBZMCHVLDF22}
Jason Wei, Yi~Tay, Rishi Bommasani, Colin Raffel, Barret Zoph, Sebastian Borgeaud, Dani Yogatama, Maarten Bosma, Denny Zhou, Donald Metzler, Ed~H. Chi, Tatsunori Hashimoto, Oriol Vinyals, Percy Liang, Jeff Dean, and William Fedus.
\newblock Emergent abilities of large language models.
\newblock {\em Trans. Mach. Learn. Res.}, 2022, 2022.

\bibitem{NEURIPS2022_9d560961}
Jason Wei and Xuezhi Wang.
\newblock Chain-of-thought prompting elicits reasoning in large language models.
\newblock In S.~Koyejo, S.~Mohamed, A.~Agarwal, D.~Belgrave, K.~Cho, and A.~Oh, editors, {\em Advances in Neural Information Processing Systems}, volume~35, pages 24824--24837. Curran Associates, Inc., 2022.

\bibitem{wolf-etal-2020-transformers}
Thomas Wolf, Lysandre Debut, Victor Sanh, Julien Chaumond, Clement Delangue, Anthony Moi, Pierric Cistac, Tim Rault, Remi Louf, Morgan Funtowicz, Joe Davison, Sam Shleifer, Patrick von Platen, Clara Ma, Yacine Jernite, Julien Plu, Canwen Xu, Teven Le~Scao, Sylvain Gugger, Mariama Drame, Quentin Lhoest, and Alexander Rush.
\newblock Transformers: State-of-the-art natural language processing.
\newblock In Qun Liu and David Schlangen, editors, {\em Proceedings of the 2020 Conference on Empirical Methods in Natural Language Processing: System Demonstrations}, pages 38--45, Online, October 2020. Association for Computational Linguistics.

\bibitem{yang-etal-2024-soft}
Jun~Cheng Yang, Zuchao Li, Shuai Xie, Wei Yu, Shijun Li, and Bo~Du.
\newblock Soft-prompting with graph-of-thought for multi-modal representation learning.
\newblock In Nicoletta Calzolari, Min-Yen Kan, Veronique Hoste, Alessandro Lenci, Sakriani Sakti, and Nianwen Xue, editors, {\em Proceedings of the 2024 Joint International Conference on Computational Linguistics, Language Resources and Evaluation (LREC-COLING 2024)}, pages 15024--15036, Torino, Italia, May 2024. ELRA and ICCL.

\bibitem{yao-etal-2024-sirllm}
Yao Yao, Zuchao Li, and Hai Zhao.
\newblock {S}ir{LLM}: Streaming infinite retentive {LLM}.
\newblock In Lun-Wei Ku, Andre Martins, and Vivek Srikumar, editors, {\em Proceedings of the 62nd Annual Meeting of the Association for Computational Linguistics (Volume 1: Long Papers)}, pages 2611--2624, Bangkok, Thailand, August 2024. Association for Computational Linguistics.

\bibitem{280922}
Gyeong-In Yu, Joo~Seong Jeong, Geon-Woo Kim, Soojeong Kim, and Byung-Gon Chun.
\newblock Orca: A distributed serving system for {Transformer-Based} generative models.
\newblock In {\em 16th USENIX Symposium on Operating Systems Design and Implementation (OSDI 22)}, pages 521--538, Carlsbad, CA, July 2022. USENIX Association.

\bibitem{zeng2022glm}
Aohan Zeng, Xiao Liu, Zhengxiao Du, Zihan Wang, Hanyu Lai, Ming Ding, Zhuoyi Yang, Yifan Xu, Wendi Zheng, Xiao Xia, Weng~Lam Tam, Zixuan Ma, Yufei Xue, Jidong Zhai, Wenguang Chen, Zhiyuan Liu, Peng Zhang, Yuxiao Dong, and Jie Tang.
\newblock {GLM-130B:} an open bilingual pre-trained model.
\newblock In {\em The Eleventh International Conference on Learning Representations, {ICLR} 2023, Kigali, Rwanda, May 1-5, 2023}. OpenReview.net, 2023.

\bibitem{zhang-etal-2024-selective}
Hongyi Zhang, Zuchao Li, Ping Wang, and Hai Zhao.
\newblock Selective prefix tuning for pre-trained language models.
\newblock In Lun-Wei Ku, Andre Martins, and Vivek Srikumar, editors, {\em Findings of the Association for Computational Linguistics ACL 2024}, pages 2806--2813, Bangkok, Thailand and virtual meeting, August 2024. Association for Computational Linguistics.

\bibitem{aclue}
Yixuan Zhang and Haonan Li.
\newblock Can large langauge model comprehend ancient chinese? {A} preliminary test on {ACLUE}.
\newblock In Adam Anderson, Shai Gordin, Bin Li, Yudong Liu, and Marco~Carlo Passarotti, editors, {\em Proceedings of the Ancient Language Processing Workshop, ALP@RANLP 2023, Varna, Bulgaria, 8 September, 2023}, pages 80--87. {INCOMA} Ltd., Shoumen, Bulgaria / {ACL}, 2023.

\bibitem{zhang2023sirens}
Yue Zhang, Yafu Li, Leyang Cui, Deng Cai, Lemao Liu, Tingchen Fu, Xinting Huang, Enbo Zhao, Yu~Zhang, Yulong Chen, Longyue Wang, Anh~Tuan Luu, Wei Bi, Freda Shi, and Shuming Shi.
\newblock Siren's song in the ai ocean: A survey on hallucination in large language models, 2023.

\bibitem{zhou2023large}
Yongchao Zhou, Andrei~Ioan Muresanu, Ziwen Han, Keiran Paster, Silviu Pitis, Harris Chan, and Jimmy Ba.
\newblock Large language models are human-level prompt engineers.
\newblock In {\em The Eleventh International Conference on Learning Representations, {ICLR} 2023, Kigali, Rwanda, May 1-5, 2023}. OpenReview.net, 2023.

\end{thebibliography}
\bibliographystyle{plain}
}

\newpage







\appendix
\section*{Appendix}



\label{sec:appendix}

\section{Testing Environments}
All testing are done on a server with 8*A100 80G SXM. For models with less than 15B parameters, 2 of 8 GPUs are used. For models with more than 15B parameters, 4 of 8 GPUs are used. All testing are carried out under HuggingFace Transformers library~\cite{wolf-etal-2020-transformers}.

\section{Generation of Reference Datastore}
\subsection{Benchmark Testing}
\label{sec:gorl}
To generate reference datastores, LLMs are shown to the questions and options in the training split of the benchmarks and we store the attention output. For each question this process is repeated four times cycling through A, B, C, D as the correct value. $3,500$ to $5,000$ question is shown to the LLMs and about $20,000$ $\left(k, v\right)$ entries are generated. To be noted is that the reference datastore of CMMLU is generated from validation set of C-eval~\cite{huang2023ceval}, split \textit{zho\_Hans} of belebele~\cite{bandarkar2023belebele} and testing set of ACLUE~\cite{aclue} since there is no training split for the benchmark.
\subsection{Wikipedia Fact Retrieval}
\label{sec:goll}
For our reference datastore, we encoded all of the Wikipedia sentences using the Jina~\cite{günther2023jina} model, which is smaller in both it's parameter count and hidden size, resulting in a faster generation speed and smaller space cost for encoded vector datastore. Every usable sentence in Wikipedia is encoded, meanwhile the sentences from the same page share a same value, which is the no. of this page. When testing, we use the same model to encode the question, then we search the most relevant pages in the datastore, be the metric of cosine similarity, to retrieve the most relevant pages.
In this section, $s_\mathcal{L}$ is same as the sentences count of Wikipedia, around $73\mathrm{M}$. $k = 1024$. $T$ and $\lambda$ are not applicable here.

With retrieved pages, we generate a reference datastore with every sentence in the pages. We first calculate attention representations for every token, whose corresponding value is the id of next token, \textit{eos} for the last token. Then we use this dynamically generated reference datastore for following RTD. In this section, $s_\mathcal{L}$ is the same as the length of tokenized sequence, $6200$ on average, and we use $T = 750, k = 1024, \lambda = 0.4$.


\section{RAG's Deficiency in Testing}
\label{sec:ragdef}
RAG method's shows a decline in performance in Table~\ref{tab:Wikipedia}. To explain this, we can further examine the average length of the tokenized sequences of the retrieved context, which is around $6200$, showcased in Table~\ref{tab:lengths}. This length will hardly increase any inference cost for the RTD method, due to the small $s_{\mathcal{L}}$, but it exceeds the pre-training sequence length of LLaMA2-7B-Chat, which is $4096$. That is to say, the naive RAG method here will cause sequence length overflow, thereby significantly affecting performance. If the overflow happened, then the model's ability is cut down significantly.


\begin{table}[b]
    \centering
    \caption{Average length by token in OBQA question answering process, split by sections.}
    \label{tab:lengths}
    \begin{tabular}{l|c}
        \toprule
        Section     & Average Length      \\
        \midrule
        Wikipedia Context   & 6192          \\
        Question            & 84      \\
        Response            & 231\\
        \bottomrule
    \end{tabular}
\end{table}

\section{LoRA Hyperparamters}
\label{sec:HPP}
See Table~\ref{tab:lorahyper}. For LoRA tuning on MMLU, any question whoes tokenized length exceed 4096 was evicted from both training and testing. The maximum tokenized length of the Tiny-Shakespeare dataset is 900. 
\begin{table}
    \centering
    \caption{LoRA Hyper-parameters}
    \label{tab:lorahyper}
    \begin{tabular}{l|c}
        \toprule
        Hyper-parameter     & Value      \\
        \midrule
        Batch Size          & 4          \\
        Epochs              & 2          \\
        Max Seq. Len.       & 4096       \\
        \midrule
        LoRA Target         & \{Q, K, V, O, Up, Down, Gate\}\_proj \\
        LoRA Rank           & 16         \\
        LoRA $\alpha$       & 32         \\
        LoRA dropout        & 0.01       \\
        \midrule
        Learning Rate       & 1e-5       \\
        Optimizer           & AdamW      \\
        Adma RMS $\epsilon$ & 2e-4       \\
        Adam $\beta$        & $(0.9, 0.999)$ \\
        Adam Weight Decay   & 0.01       \\
        Scheduler           & Constant LR\\
        \bottomrule
    \end{tabular}
\end{table}


\newpage
\section*{NeurIPS Paper Checklist}

\begin{enumerate}

\item {\bf Claims}
    \item[] Question: Do the main claims made in the abstract and introduction accurately reflect the paper's contributions and scope?
    \item[] Answer: \answerYes{} 
    \item[] Justification: Section~\ref{sec:RTDMethod} and section~\ref{sec:exper} reflects our main claim.

\item {\bf Limitations}
    \item[] Question: Does the paper discuss the limitations of the work performed by the authors?
    \item[] Answer: \answerYes{} 
    \item[] Justification: Section~\ref{sec:limits}.

\item {\bf Theory Assumptions and Proofs}
    \item[] Question: For each theoretical result, does the paper provide the full set of assumptions and a complete (and correct) proof?
    \item[] Answer: \answerYes{} 
    \item[] Justification: Our main theoretical result is about efficiencies in section~\ref{sec:effie}, in which they were proved.

    \item {\bf Experimental Result Reproducibility}
    \item[] Question: Does the paper fully disclose all the information needed to reproduce the main experimental results of the paper to the extent that it affects the main claims and/or conclusions of the paper (regardless of whether the code and data are provided or not)?
    \item[] Answer: \answerYes{} 
    \item[] Justification: The detailed description of our experiments and hyper-parameters can be found in section~\ref{sec:exper} and appendix~\ref{sec:HPP}.

\item {\bf Open access to data and code}
    \item[] Question: Does the paper provide open access to the data and code, with sufficient instructions to faithfully reproduce the main experimental results, as described in supplemental material?
    \item[] Answer: \answerYes{} 
    \item[] Justification: Our main codes can be found in Supplementary Material.

\item {\bf Experimental Setting/Details}
    \item[] Question: Does the paper specify all the training and test details (e.g., data splits, hyperparameters, how they were chosen, type of optimizer, etc.) necessary to understand the results?
    \item[] Answer: \answerYes{} 
    \item[] Justification: Discussions of hyper-parameters of our methods can be found in section~\ref{sec:ihpr}. 

\item {\bf Experiment Statistical Significance}
    \item[] Question: Does the paper report error bars suitably and correctly defined or other appropriate information about the statistical significance of the experiments?
    \item[] Answer: \answerNo{} 
    \item[] Justification: The experimental results are definitive and do not involve any random factors. 

\item {\bf Experiments Compute Resources}
    \item[] Question: For each experiment, does the paper provide sufficient information on the computer resources (type of compute workers, memory, time of execution) needed to reproduce the experiments?
    \item[] Answer: \answerYes{} 
    \item[] Justification: All information above are given in section~\ref{sec:exper} and with specific experiment focused on some of them.
    
\item {\bf Code Of Ethics}
    \item[] Question: Does the research conducted in the paper conform, in every respect, with the NeurIPS Code of Ethics \url{https://neurips.cc/public/EthicsGuidelines}?
    \item[] Answer: \answerYes{} 
    \item[] Justification: \answerNA{}

\item {\bf Broader Impacts}
    \item[] Question: Does the paper discuss both potential positive societal impacts and negative societal impacts of the work performed?
    \item[] Answer: \answerNA{} 
    \item[] Justification: There is no societal impact of the work performed.
    
\item {\bf Safeguards}
    \item[] Question: Does the paper describe safeguards that have been put in place for responsible release of data or models that have a high risk for misuse (e.g., pretrained language models, image generators, or scraped datasets)?
    \item[] Answer: \answerNA{} 
    \item[] Justification: The paper poses no such risks.

\item {\bf Licenses for existing assets}
    \item[] Question: Are the creators or original owners of assets (e.g., code, data, models), used in the paper, properly credited and are the license and terms of use explicitly mentioned and properly respected?
    \item[] Answer: \answerYes{} 
    \item[] Justification: All models and datasets we've used have been cited properly in section~\ref{sec:exper}. 

\item {\bf New Assets}
    \item[] Question: Are new assets introduced in the paper well documented and is the documentation provided alongside the assets?
    \item[] Answer: \answerYes{} 
    \item[] Justification: Our main codes, as assets, are updated in Supplementary Material.

\item {\bf Crowdsourcing and Research with Human Subjects}
    \item[] Question: For crowdsourcing experiments and research with human subjects, does the paper include the full text of instructions given to participants and screenshots, if applicable, as well as details about compensation (if any)? 
    \item[] Answer: \answerNA{} 
    \item[] Justification: The paper does not involve crowdsourcing nor research with human subjects.

\item {\bf Institutional Review Board (IRB) Approvals or Equivalent for Research with Human Subjects}
    \item[] Question: Does the paper describe potential risks incurred by study participants, whether such risks were disclosed to the subjects, and whether Institutional Review Board (IRB) approvals (or an equivalent approval/review based on the requirements of your country or institution) were obtained?
    \item[] Answer: \answerNA{} 
    \item[] Justification: The paper does not involve crowdsourcing nor research with human subjects.

\end{enumerate}

\end{document}